\documentclass[conference]{IEEEtran}
\IEEEoverridecommandlockouts
\usepackage{cite}
\usepackage{amsmath,amssymb,amsfonts}
\usepackage{algpseudocode}
\usepackage{graphicx}
\usepackage{textcomp}
\usepackage{xcolor}
\usepackage{algorithm}
\usepackage{algpseudocode}
\usepackage{booktabs}
\usepackage{soul}

\def\BibTeX{{\rm B\kern-.05em{\sc i\kern-.025em b}\kern-.08em
    T\kern-.1667em\lower.7ex\hbox{E}\kern-.125emX}}
\begin{document}

\title{Cooperative Robotics Reinforced by Collective Perception for Traffic Moderation\\
}

\author{
  \IEEEauthorblockN{
    Mohammad Khoshkdahan\IEEEauthorrefmark{1},
    John Pravin Arockiasamy\IEEEauthorrefmark{1},
    Andy Flores Comeca\IEEEauthorrefmark{1},
    Alexey Vinel\IEEEauthorrefmark{1}\IEEEauthorrefmark{2}
  }

  \IEEEauthorblockA{
    \IEEEauthorrefmark{1}Karlsruhe Institute of Technology, 
    Karlsruhe, Germany \\
    Email: alexey.vinel@kit.edu
  }

  \IEEEauthorblockA{
    \IEEEauthorrefmark{2}Halmstad University, Halmstad, Sweden
    \\
    Email: alexey.vinel@hh.se
  }

  \thanks{\IEEEauthorrefmark{1}\,The first three authors contributed equally to this work.}
}

\maketitle

\begin{abstract}
Collisions at non-line-of-sight (NLOS) intersections remain a major safety concern because drivers have limited visibility of approaching traffic. V2X based warnings can reduce these risks, yet many vehicles are not equipped with V2X and drivers may ignore in vehicle alerts. Collective perception (CP) can compensate for low V2X penetration by extending the awareness of connected vehicles, but it cannot influence unconnected vehicles. To fill this gap, our work introduces a complementary concept that adds a cooperative humanoid robot as an active traffic moderator capable of physically stopping a vehicle that attempts to merge into an unseen traffic stream.

The system operates on two parallel perception pathways. A dual camera infrastructure unit detects the position, speed and motion of approaching vehicles and transmits this information to the robot as a collective perception message (CPM). The robot also receives cooperative awareness messages (CAM) from connected vehicles through its onboard V2X unit and can act as a relay for decentralized environmental notification messages (DENM) when safety events originate elsewhere along the road. A fusion module combines these streams to maintain a robust real time view of the main road. A Zone of Danger (ZoD) is defined and used to predict whether an approaching vehicle creates a collision risk for a merging road user. When such a risk is detected, the robot issues a human-like STOP gesture and blocks the merging path until the hazard disappears.

The full system was deployed at the Future Mobility Park (FMP) in Rotterdam. Experiments show that the combined vision and V2X perception allows the robot to detect approaching vehicles early, predict hazards reliably and prevent unsafe merges in real world NLOS conditions.
\end{abstract}


\section{Introduction}
High traffic volumes and increasingly complex road layouts require drivers to maintain strong awareness of their surroundings. Among these, intersections or merging points where a local road joins a main road with limited visibility present particular danger. In such non-line-of-sight (NLOS) scenarios, drivers on both roads may not see each other until it is too late. According to the U.S. Federal Highway Administration, intersections account for about 40 percent of all crashes and roughly 22 percent of fatal crashes worldwide \cite{harkey2021impact}.

NLOS situations are especially critical in areas without traffic lights or priority signs, such as exits from campuses, private facilities, or parking areas that connect directly to a main road. In these cases, a curb side marking is often the only indication of lower priority, while the view is blocked by parked cars, buildings, or temporary construction barriers. Naturalistic driving studies report mean brake reaction times of approximately 1.5~s with substantial variance \cite{arbabzadeh2019hybrid}, and controlled experiments show that reaction delays grow under reduced visibility \cite{schlurscheid2024analysis}, which reduces the available response window in NLOS merging scenarios to well under two seconds.
 Even modern vehicles with advanced sensors cannot detect cross traffic that remains fully hidden or approaches from a sharp angle, leading to frequent near misses and side collisions.

To address these limitations, we propose a robotic traffic moderation system based on collective perception \cite{delooz2022analysis}. A vision based infrastructure setup monitors the main road, while a mobile robot positioned at the local road gate observes vehicles exiting. Both units exchange real time traffic information through a Vision-V2X fusion framework, which allows the robot to interpret approaching traffic even when the line of sight is blocked and to guide vehicles safely with police-like gestures.

\begin{figure}[t]
    \centering
    \includegraphics[width=0.96\linewidth]{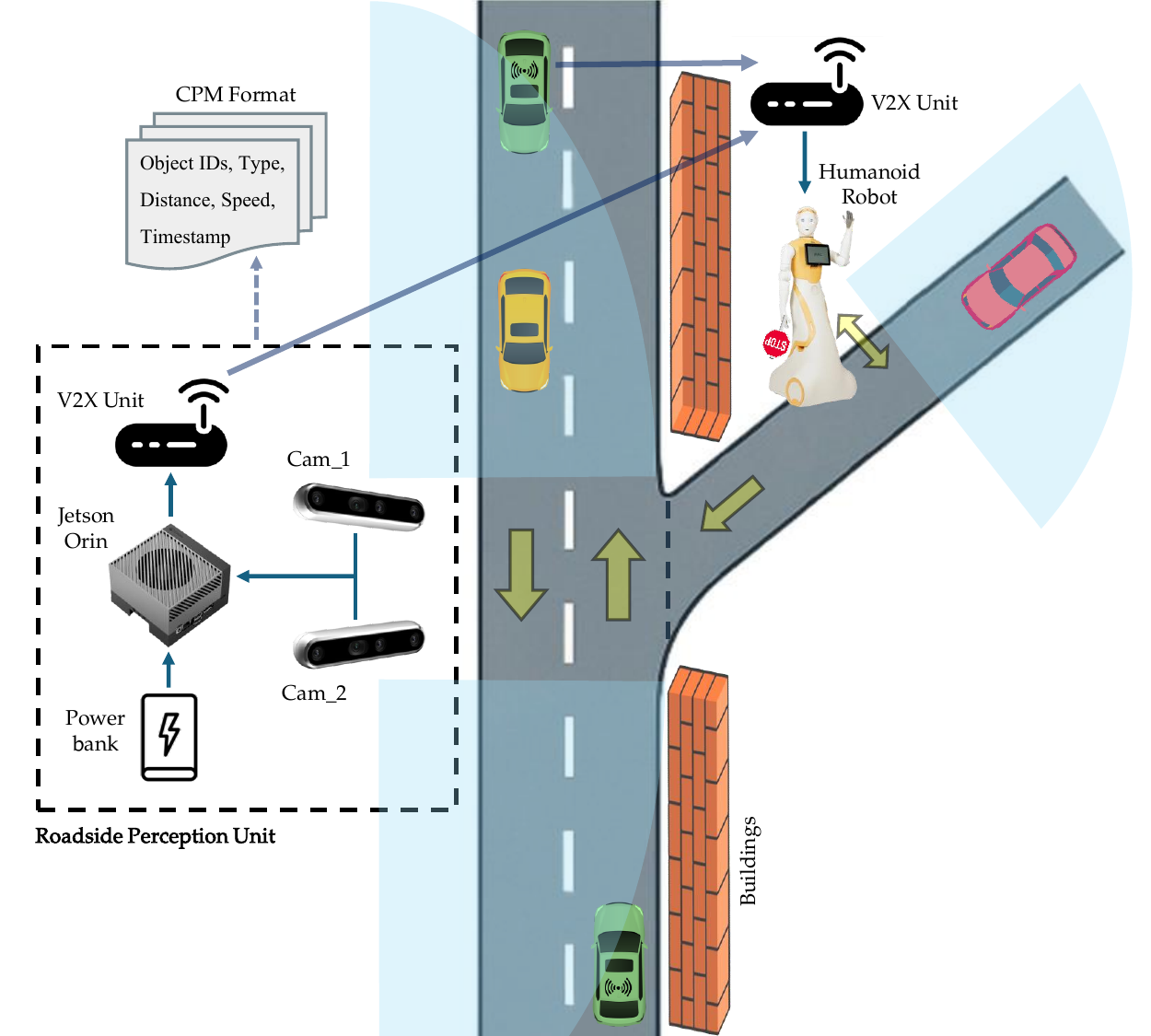}
    \caption{The overview of the proposed system in a NLOS road merging.}
    \label{fig:overview}
\end{figure}

Our main contribution is the design of a cooperative system that uses a humanoid robot, named ARI from Pal Robotics\footnote{https://pal-robotics.com/robot/ari/}, as an autonomous traffic moderator. The robot is equipped with a V2X communication unit which supports the set of ETSI messages such as Cooperative Awareness Message (CAM)~\cite{ETSI-CAM} for periodic status broadcasting, Collective Perception Message (CPM)~\cite{ETSI-CPM} for cooperative perception sharing of detected objects and sensor data, and Decentralized Environmental Notification Message (DENM)~\cite{ETSI-DENM} for event-triggered hazard and warning notification. As an autonomous traffic moderator, the robot can move to the center of the road to intervene when a potential collision is detected, using hand gestures to stop a vehicle attempting to enter the main road. Through CPM, the robot receives real time information about traffic conditions on the main road from a dual camera infrastructure system installed near the intersection. In addition, it receives CAM from connected vehicles within the communication range. By combining the infrastructure based CPM stream with the CAM information from vehicles, the robot attains a more reliable understanding of the environment and makes accurate decisions about when it is safe for a vehicle to proceed. As an additional capability, the robot can rebroadcast DENM (e.g., those transmitted by nearby safety vehicles), thereby extending the hazard notification range beyond the original communication range of the transmitting vehicle. The complete system has been integrated and tested in real world conditions at the Future Mobility Park (FMP) in the Netherlands.

This paper is organized as follows. Section~\ref{sec:rl} reviews related work on traffic moderation and collision avoidance. The next sections introduce the infrastructure sensing unit and V2X components together with the robot moderator and present the detailed logic used for collision prediction and intervention.

\section{Related Works} \label{sec:rl}
\subsection{Infrastructure-Assisted Perception}

Several studies have investigated the role of infrastructure-based perception systems for traffic safety and collision avoidance.  
This work \cite{anisha2023automated} fused camera and LiDAR data using YOLO-v5 and Kalman tracking to detect vehicle-to-vehicle conflicts at signalised intersections. However, their evaluation was limited to intersections with good visibility and line-of-sight, and did not address NLOS conditions. While multi-sensor fusion can improve robustness, it also increases computational complexity, which may be less suitable for mobile platforms with limited onboard resources. In such cases, single-sensor perception can be a more efficient alternative. Recent work has shown that real-time 3D object detection in urban environments is feasible using LiDAR-only perception \cite{khoshkdahan2026TriBand}. Camera-only systems also remain attractive due to their low cost and practical deployment, although fairness-related issues in pedestrian detection should be taken into account \cite{khoshkdahan2025fair,khoshkdahan2025beyond}.
Another work \cite{wu2023multi} presented a video and LiDAR analytics system for pedestrian–vehicle interaction analysis at intersections, focusing primarily on vulnerable road users rather than dynamic merging traffic flows.
The study in \cite{liu2018cooperation} explored a hybrid perception system combining roadside radar with V2I/P2I communication to detect occluded vulnerable road users, thereby allowing detection beyond direct line-of-sight.
The work \cite{zhang2021longitudinal} proposed fusing traffic video with static 3D LiDAR infrastructure data to reconstruct trajectories of road users, supporting proactive safety monitoring of vehicle behaviour.
While these works advance infrastructure-based sensing and cooperative perception, they assume clear visibility, emphasise pedestrian safety, or focus on trajectory reconstruction. Our system fills the gap by combining roadside vision sensing, V2X data exchange, and a robotic moderator to directly manage merging traffic in NLOS situations.

\subsection{V2X-Based Traffic Management}

Several studies use V2X communication to improve safety at intersections and in limited-visibility scenarios. The distributed right-of-way algorithm in AROW  \cite{shah2024arow} assigns explicit crossing turns among connected vehicles at unsignalized intersections and resolves priority ambiguity via V2X message exchange. A vector-based forward collision warning scheme (V-FCW) transmits position, velocity, and heading via V2V/V2I and uses map information to reliably detect hazards even on curved or non-standard roads \cite{Cai2025V}. Hardware-in-the-loop tests showed that V2X-based intersection warning systems can drastically reduce collision risk under realistic radio-channel conditions \cite{bazzi2020hardware}. The cooperative collision avoidance method introduced in \cite{gelbal2023cooperative} predicts future trajectories from V2X data and resolves imminent conflicts at obstructed intersections, even when onboard sensors cannot see the hazard. It demonstrates effective avoidance in NLOS scenarios with limited reaction time.

In the state-of-the-art V2X systems the main remedy for low penetration is collective perception, which helps connected vehicles detect non-equipped users but cannot actively influence them. Existing approaches therefore remain limited to warnings or automated control inside the equipped vehicles themselves. Even recent eHMI and V2P solutions that utilize personal devices to deliver visual or vibrotactile warnings provide only communicative signals rather than physical actuation \cite{lanzer2023interaction}. Our work addresses this gap by combining V2X and infrastructure-vision data with a robotic traffic moderator that can issue explicit commands and physically intervene, creating a way to incorporate non-V2X-equipped users into the cooperative driving ecosystem.

\subsection{Traffic Moderation Using Robots}

Humanoid police officers for traffic moderation are still rarely explored. Most research remains at the simulation level. This work \cite{najjar2022towards} introduced a robot model for managing intersections through hand signals in a web based simulator, but it was never tested in real traffic and did not involve interaction with actual vehicles.

Another study \cite{ghaffar2022controlling} used a TIAGO-based humanoid robot designed to act as a traffic officer with scripted gestures. Although it demonstrated gesture recognition and social acceptance, the experiment was limited to a controlled setting without real vehicles or dynamic traffic flows.

The system proposed in \cite{comeca2025robots, comeca2025social} is designed to explore humanoid robots for assisting pedestrian crossings. These systems demonstrate that physical gestures can support pedestrian safety, yet they focus solely on pedestrian interaction and do not address vehicle-to-vehicle hazards. Without infrastructure supported perception their field of view remains limited, which prevents their application in NLOS merging scenarios.

Unmanned aerial vehicles have been studied mainly for traffic observation. Recent UAV-based systems integrate onboard cameras and wireless networks to monitor traffic density, detect near-miss events, and assess safety through cooperative sensing \cite{mahendran2025design, masuduzzaman2022uav, butilua2022urban}. While effective for surveillance and data collection, UAVs remain passive observers that cannot directly influence vehicle drivers. Their operation is also limited by occlusion in urban areas, short endurance, and the difficulty of maneuvering in complex environments.

\section{Traffic Moderator Robot}

\begin{figure}[t]
    \centering
    \includegraphics[width=1\linewidth]{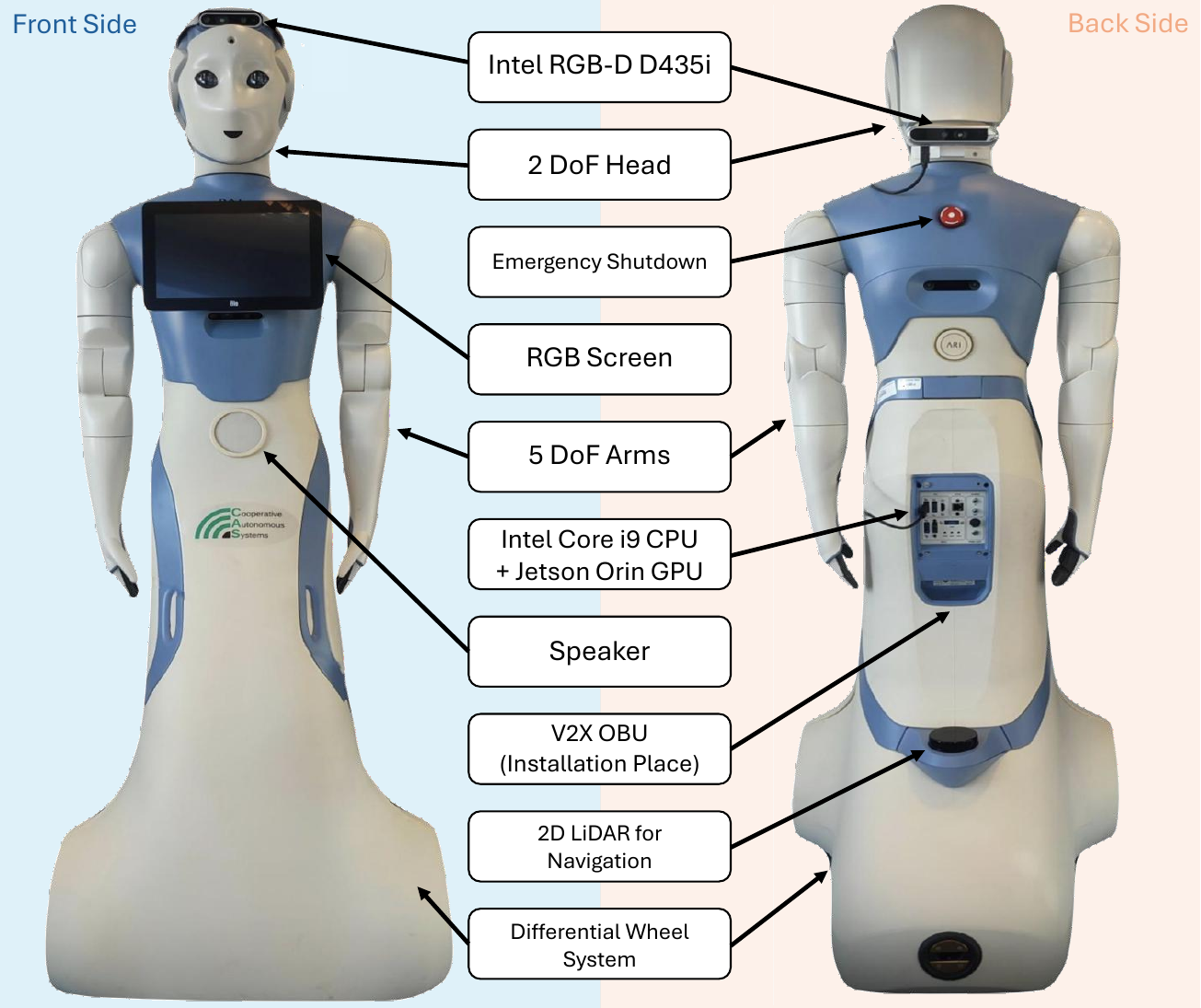}
    \caption{Front and rear views of the ARI robot with its sensing, computation and actuation modules.}
    \label{fig:ari-robot}
\end{figure}

ARI (shown in Fig. \ref{fig:ari-robot}) is a mobile humanoid robot designed for autonomous operation. It carries a front-mounted RGB-D camera based on the Intel RealSense D435, which provides color and depth information for detecting nearby vehicles, cyclists and pedestrians. A 2D LiDAR supports SLAM based localization and mapping so the robot can navigate safely around the FMP gate where the side road merges with the main road without colliding with obstacles. A V2X communication unit is mounted on the back of the robot and provides direct access to wireless traffic information during deployment.

For perception we use a lightweight YOLOv12 \cite{tian2025yolov12} network that identifies approaching vehicles and cyclists from the head camera of ARI. The robot’s head has two degrees of freedom (DoF) to extend its field of view and he is using a differential-drive base for moving. Its two five-DoF arms perform the \textit{STOP} and \textit{PASS} gestures required for traffic moderation. The robot is equipped with an Intel Core i9 processor for navigation and motion control together with a Jetson Orin AGX GPU for real-time vision inference.

\section{Infrastructure Perception System}
\subsection{Hardware Setup}
The infrastructure perception system consists of two Intel D455 cameras mounted on a pole at a height of 1.5 meters. The depth information is not used, as the available range is limited. Image processing is handled by a Jetson AGX Orin powered by a portable power bank. The system periodically transmits mandatory CPM fields, including \textit{sensorInformation} and \textit{perceivedObject} attributes (objectId, position, velocity, and ObjectClass), to the robot via the V2X unit at 5 Hz.

\subsection{Vehicle Detection and Parameter Estimation}

\begin{figure}
    \centering
    \includegraphics[width=1\linewidth]{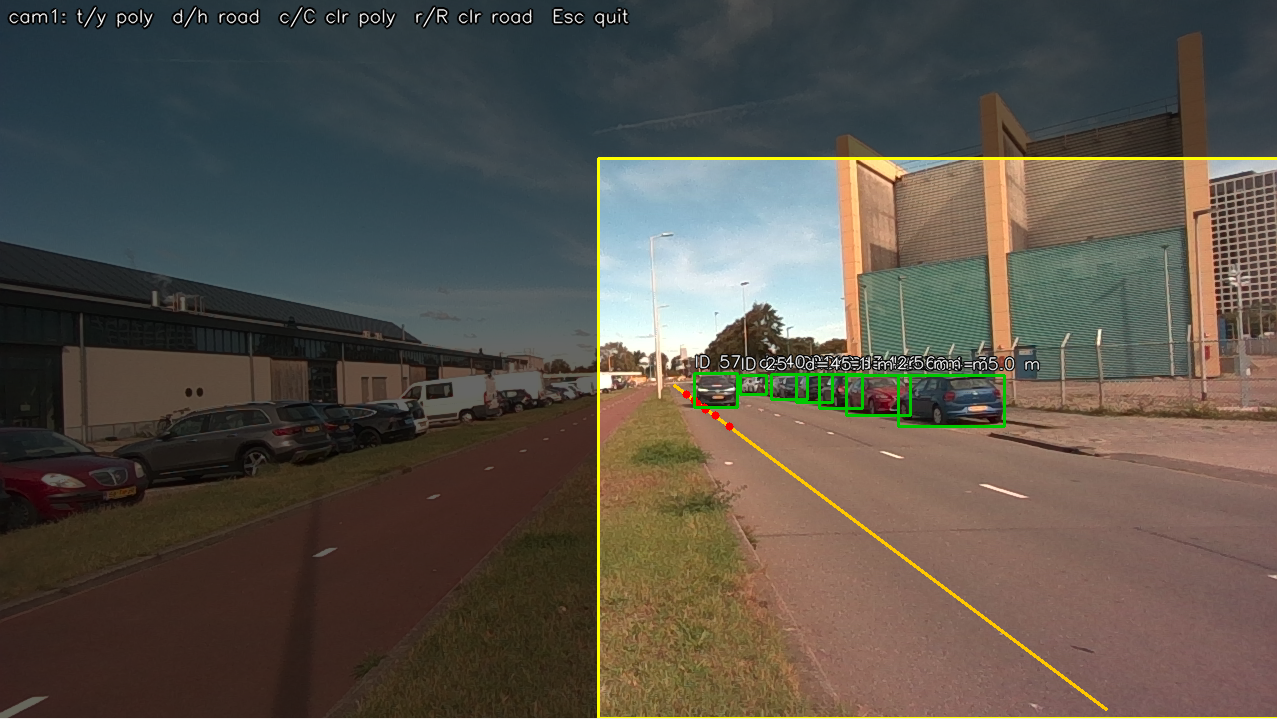}
    \caption{Detection output of the dual camera infrastructure system showing vehicle tracking and projection of bounding boxes onto the road reference line.}
    \label{fig:cam_det}
\end{figure}

We use the robot operating system (ROS) to acquire image frames and to assemble the mandatory CPM fields. Each camera stream is fed to a pretrained YOLOv12 (N-variant) detector \cite{tian2025yolov12} with approximately 2.6 million parameters, chosen for its short inference time suitable for real-time operation. Detections are assigned unique track identifiers, and the 2D bounding box information is retrieved for each frame as illustrated in Fig. \ref{fig:cam_det}. Since vehicles only move along a single main road, each frame is cropped to the road region with a small surrounding margin, which reduces input dimensions and achieves an average reduction of 55.2\% in inference time.

For distance estimation, the road is first calibrated. An imaginary straight reference line is overlaid along the lane center in the image frame, parameterized by a linear coordinate \( s \in [0,S] \) from the camera origin. A set of calibration pairs \( \mathcal{C}=\{(s_i,d_i)\}_{i=1}^{N} \) is obtained by placing a test vehicle at ten known positions up to 120~m from the camera origin and recording the true distance \( d_i \) using GPS. Given a detection with bottom-center pixel \( b_t \), its horizontal projection onto the line is \( s_t = \Pi_{\text{line}}(b_t) \). The distance is then computed through a polynomial regression function

\begin{equation}
\hat d_t = f(s_t) = \mathbf{w}^\mathsf{T}\boldsymbol{\phi}(s_t)
\end{equation}

where \( \boldsymbol{\phi}(s_t) = [1,\,s_t,\,s_t^2,\ldots,s_t^k]^\mathsf{T} \) is the polynomial feature vector and \( \mathbf{w} \) is the regression coefficient vector estimated from the calibration data. Let the design matrix be 
\( \boldsymbol{\Phi} = [\boldsymbol{\phi}(s_1), \boldsymbol{\phi}(s_2), \ldots, \boldsymbol{\phi}(s_N)]^\mathsf{T} \)
and \( \mathbf{d} = [d_1, d_2, \ldots, d_N]^\mathsf{T} \) the corresponding distance vector. The optimal weights are obtained using least squares as

\begin{equation}
\mathbf{w} = (\boldsymbol{\Phi}^\mathsf{T}\boldsymbol{\Phi})^{-1}\boldsymbol{\Phi}^\mathsf{T}\mathbf{d}.
\end{equation}

A low polynomial order \( k\!\leq\!2 \) provides a good trade-off between smoothness and accuracy, since calibration points are closely spaced along the line.

For motion and speed estimation, each tracked object is observed at three consecutive timestamps \( t_1 < t_2 < t_3 \) with corresponding estimated distances \( \hat d_{t_1}, \hat d_{t_2}, \hat d_{t_3} \). The average velocity is computed as

\begin{equation}
\bar v = \frac{1}{2}\!\left(
\frac{\hat d_{t_2}-\hat d_{t_1}}{t_2-t_1} +
\frac{\hat d_{t_3}-\hat d_{t_2}}{t_3-t_2}
\right)
\end{equation}

An object is labeled as moving when \( |\bar v| > 3\,\text{m/s} \); the sign of \( \bar v \) indicates whether the object is approaching or receding relative to the camera. This threshold filters out small apparent motions caused by camera vibration or minor bounding-box variations on stationary vehicles due to model uncertainty.

All estimated parameters, including object ID, class, distance, speed, and timestamp, are packed into a single CPM and transmitted at 5~Hz through the V2X unit to the robot.

\section{V2X Communication and Data Exchange}
\subsection{V2X Hardware and Protocol}

The system uses three ETSI-compliant On-Board Units (OBUs), each operating over IEEE 802.11p (ITS-G5) based Dedicated Short-Range Communication (DSRC). A V2X-equipped Volkswagen (VW) ID.4 transmits CAMs. The infrastructure perception system publishes CPMs through a Herman OBU\footnote{https://www.herman.cz/en/}. ARI carries a second Herman OBU that broadcasts CAMs using the ETSI pedestrian profile as no official robot-specific profile currently exists. A safety car with a Roadside Unit (RSU) sends DENM messages (e.g., with causecode "roadworks") to emulate safety-related event triggers.  ARI receives CAMs from the VW ID.4, CPMs from the infrastructure perception system, and DENMs from the RSU.


The ITS-G5 stack provides low-latency vehicle-to-infrastructure as well as vehicle-to-vehicle messaging. In our implementation, the OBU interfaces with ROS via a custom bridge node, which is an integration between the V2X data stream and the robot’s perception module.

\subsection{Hazard Detection}

CAMs from V2X-equipped vehicles and CPMs from the infrastructure cameras (detection results) are merged in the robot’s perception module. Each input provides object position along the road reference line together with heading and timestamp, and camera-only detections are mapped onto this line through the calibrated projection.

Because ARI observes the main road from the side rather than along the vehicle trajectory, Time to Collision (ToC) becomes less reliable in this geometry \cite{banjade2021vulnerable}. We therefore use a region based approach, referred to as the Zone of Danger (ZoD), that predicts whether an approaching vehicle will enter a predefined conflict area. The ZoD is dimensioned to cover the stopping distance of a fast approaching vehicle at up to \(80\,\mathrm{km/h}\), including both braking distance and typical driver reaction time, and to match the layout of the FMP. This results in a \(50\times50\,\mathrm{m}\) area around the robot. The temporal threshold for intervention is set to \(\tau_{\mathrm{th}} = 5\,\mathrm{s}\), which matches the time ARI requires to move from its standby position to the centre of the lane and raise the STOP gesture, with only a minor portion attributed to perception and decision processing.

For each detected object the system estimates the entry time \(t_{\mathrm{enter}}\) into the ZoD and the exit time \(t_{\mathrm{exit}}\). If \(t_{\mathrm{enter}} \le \tau_{\mathrm{th}}\) the robot issues a STOP gesture and blocks the merging movement until the object leaves the ZoD. Otherwise merging is permitted.

\subsection{Scenario-Level and Communication-Level KPIs} \label{sec:KPI}

To evaluate system behavior during each demonstration run, we define several scenario-level key performance indicators (KPIs). For the test vehicle (VW ID.4), we record the total duration of the run, the total distance traveled, and compute its average speed. We also measure the duration the test vehicle remains within the predefined ZoD region, referred to as VW ZoD Time. To assess the influence of the robot moderator, we log the total time during which ARI blocked merging traffic (ARI Stop Time), based on the five-second threshold for ZoD entry.

On the communication side, we evaluate ARI’s inter-generation gap (IGG), defined as the average interval between consecutive CAM generated by ARI to inform other road users about its presence. We also monitor the inter-packet gap (IPG), defined as the average interval between reception of consecutive CAM packets received from a VW ID.4 to check the regularity of incoming messages. Furthermore, we evaluate the latency between the infrastructure system sending the CPM and ARI receiving it, referred further as CPM Latency. These metrics together indicate how reliable and timely the V2X channel is, which is critical for real-time perception fusion and robot decision making.

\section{Framework Logic}

\begin{figure}[t]
    \centering
    \includegraphics[width=1\linewidth]{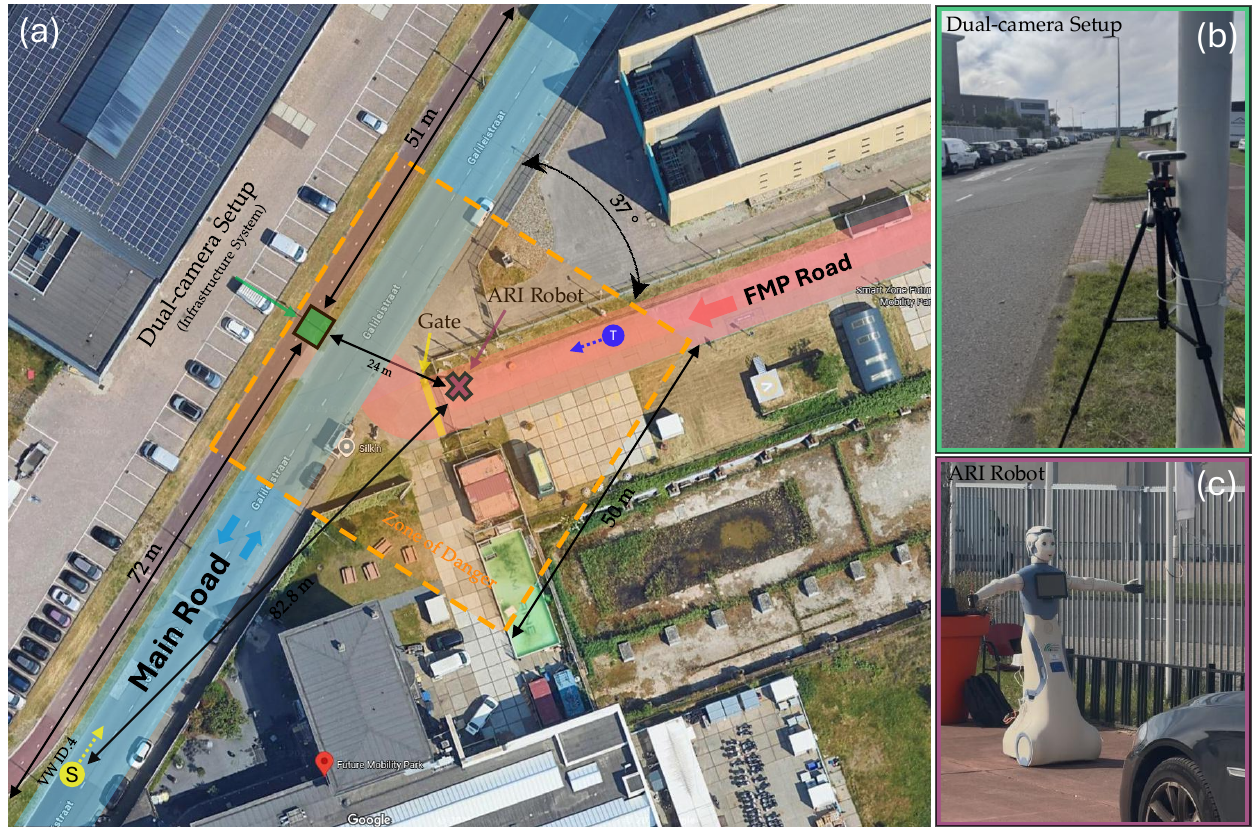}
\caption{Experimental setup at the FMP.  
(a) Test site map showing the main road, the FMP road, the merging geometry and the defined ZoD. In this example, the V2X-equipped VW~ID.4 (yellow) approaches along the main road while a test vehicle inside the FMP (blue) attempts to enter it.  
(b) Dual-camera infrastructure system used for long-range vehicle detection on the main road.  
(c) ARI robot executing the STOP gesture while moderating traffic at the FMP gate.}
    \label{fig:fmp_map}
\end{figure}

The proposed system was implemented and tested at the FMP in Rotterdam, Netherlands. As illustrated in Fig.~\ref{fig:fmp_map}, the red path represents the narrow internal road used by vehicles and cyclists exiting toward the main road. The ARI humanoid robot is positioned at the gate to monitor and regulate the internal traffic flow. It only grants passage when no potential collision is detected. Possible collisions are predicted from two input sources: (1) the roadside camera system detecting approaching vehicles from both directions of the main road, and (2) V2X communication data received from vehicles equipped with on-board units (OBUs). The detailed decision flow that combines these perception streams is formalised in Algorithm~\ref{alg:robot_logic}. To avoid tracking the same vehicle twice the logic uses a priority based fusion method. It trusts V2X data first and only includes camera detections if they are far enough away from known V2X vehicles. For each tracked vehicle the system calculates when it will enter and leave the Zone of Danger. If a vehicle will enter this zone within a short time limit while the robot sees a local user trying to merge a hazard is triggered. The robot then steps in to block the lane and maintains the stop command until the vehicle clears the area.

\begin{algorithm}[t]
\caption{Cooperative Vision–V2X Traffic Moderation Algorithm (ZoD + V2X-priority + Local-Merge Check)}
\label{alg:robot_logic}
\begin{algorithmic}[1]
\Require Infrastructure camera CPMs \(\mathcal{M}_{c,L}, \mathcal{M}_{c,R}\), V2X messages \(\mathcal{M}_v\) from road vehicles, local robot-camera feed \(\mathcal{M}_r\)
\Ensure Safe merging control at NLOS intersection
\State \(\mathcal{S} \gets \textsf{SAFE}\)
\While{system active}
    \State Receive \(\mathcal{M}_{c,L}(t), \mathcal{M}_{c,R}(t), \mathcal{M}_v(t), \mathcal{M}_r(t)\)
    \State Parse \(\mathcal{O}_v(t)\) from V2X and \(\mathcal{O}_c(t)\) from camera detections
    \State Project all object positions onto the road reference line
    \State \(\mathcal{O}_f(t) \gets \mathcal{O}_v(t)\)  \Comment{Prioritise V2X detections}
    \ForAll{\(o_j \in \mathcal{O}_c(t)\)}
        \If{\(\min_{o_i \in \mathcal{O}_v(t)} \|(x_j,v_j)-(x_i,v_i)\| \ge \epsilon\)}
            \State \(\mathcal{O}_f(t) \gets \mathcal{O}_f(t) \cup \{o_j\}\)
        \EndIf
    \EndFor
    \ForAll{\(o_i \in \mathcal{O}_f(t)\)}
        \State Estimate \(t_{\mathrm{enter},i}\) and \(t_{\mathrm{exit},i}\) for ZoD entry/exit  
    \EndFor
    \If{\(\exists\, o_k \in \mathcal{O}_f(t)\) such that \(t_{\mathrm{enter},k} \le \tau_{\mathrm{th}}\) and robot camera \(\mathcal{M}_r\) sees a merging vehicle}
        \State \(\mathcal{S} \gets \textsf{DANGER}\)
        \State Issue \(\mathbf{STOP}\) gesture: move robot to lane center, show stop sign
        \While{object \(o_k\) remains inside ZoD (i.e. current time \(< t_{\mathrm{exit},k}\))}
            \State Maintain STOP posture
            \State Receive and process new data streams
            \State Update object states and recalculate \(t_{\mathrm{exit},k}\)
        \EndWhile
        \State \(\mathcal{S} \gets \textsf{SAFE}\)
    \Else  
        \If{\(\mathcal{S} = \textsf{SAFE}\)}  
            \State Issue \(\mathbf{PASS}\) gesture: arms to neutral; step back to clear lane centre  
        \EndIf
    \EndIf
    \State Log \(\mathcal{L}(t) \gets (\mathcal{O}_f(t), \mathcal{S}, \{t_{\mathrm{enter},i}, t_{\mathrm{exit},i}\})\)
\EndWhile
\end{algorithmic}
\end{algorithm}

\section{Experiments and Results}

\subsection{Main Experiments at FMP}
The complete system was evaluated during a four hour deployment in the FMP on an active main road with mixed vehicle traffic (see Fig. \ref{fig:fmp_map}). To test the fusion of infrastructure sensing and V2X, a VW ID.4 equipped with a V2X unit performed repeated passes along the monitored segment while the robotic moderator remained stationed at the merging gate. 

During the full deployment, 798 approaching vehicles were recorded. The camera-based infrastructure system, positioned 24 m from ARI, detected all vehicles on the main road, with an average first detection distance of 110.1 m and a standard deviation of 6.5 m. Large trucks and buses showed slightly shorter initial detection distances, but remained within a range that allowed sufficient time for hazard assessment.

For the V2X experiment, the VW ID.4 started at 82.8 m and stopped at 127 m (on the other end of the road) relative to ARI. The recorded V2X messages provided a continuous trajectory of the vehicle. The VW ID.4 was tracked for a total duration of 19.82\,s, from the beginning of the sequence until it came to a complete stop. During this interval, the vehicle maintained an average speed of 9.35\,m/s and covered approximately 185.5\,m. From the trajectory, we extracted the time the VW spent inside the ZoD (VW ZoD Time) and the corresponding duration during which ARI maintained the STOP gesture (ARI Stop Time). The relationship between these values (illustrated in Table~\ref{tab:kpis}) shows that ARI consistently initiated intervention before the vehicle reached the boundary of ZoD and released the road shortly after it exited, ensuring that the side road was not blocked longer than necessary.

Communication behavior was analyzed using the timing indicators defined in Section~\ref{sec:KPI}. Table~\ref{tab:kpis} summarizes all scenario-level and communication-level KPIs measured during the 19.82 s tracking interval. Despite complete NLOS between ARI, the vehicle, and the infrastructure perception system, the observed message timing remained regular and no major degradation of the V2X link occurred. Since ARI remained stationary, CAMs were generated solely according to the ETSI maximum generation interval~\cite{ETSI-CAM}, resulting in an IGG of approximately 1\,s. CAM reception from the vehicle was stable, and CPM arrivals from the infrastructure cameras were sufficiently frequent to maintain an accurate view of main road traffic.

\begin{table}[t]
\centering
\caption{Scenario-level and communication KPIs from the Rotterdam experiment. All values are in seconds.}
\label{tab:kpis}
\renewcommand{\arraystretch}{1.1}
\setlength{\tabcolsep}{20pt}
\begin{tabular}{lc}
\toprule
Metric & Value \\
\midrule
ARI IGG         & 1.10 \\
VW IPG          & 0.50 \\
CPM Latency     & 1.30 \\
VW ZoD Time     & 10.99 \\
ARI STOP Time   & 12.59 \\
\bottomrule
\end{tabular}
\end{table}

\subsection{DENM Repeater Functionality as a System Extension}

Although the primary aim of this work is collision avoidance using vision and V2X fusion, the presence of a mobile V2X-enabled robot enables an additional extension in which event-driven DENM are boosted to increase the initial range. In this scenario, a safety car with an RSU positioned at the end of the main road broadcasts DENM messages describing safety-related events. ARI, located on the FMP road approximately 134.3\,m from the RSU, can receive these messages, whereas the VW ID.4 at the starting point cannot, due to the greater distance (approximately 220\,m from the RSU). ARI then rebroadcasts the received DENMs, making them immediately available to the VW ID.4, which is 82.8\,m from ARI. This allows the vehicle to decode the forwarded message even when the original transmitter is not yet reachable, effectively enabling ARI to function as a DENM repeater. As the vehicle moves along the road, it subsequently receives the same DENM from both ARI and the RSU. Additionally, the data stream from the RSU remains stable throughout the experiment, with an average inter-packet gap of \(0.522\,\mathrm{s}\), demonstrating reliable message reception by ARI.

By acting as a mobile relay, ARI expands the spatial coverage of hazard notifications and mitigates communication blind spots in NLOS conditions.

\section{Discussion and Future Directions}

The system predicted all hazardous situations and the robot successfully stopped the merging vehicle in every case. The dual perception paths ensured consistent situational awareness, yet several improvements can further enhance performance.

The detection range for non-V2X vehicles is limited by camera resolution and road geometry. Adding a long-range radar sensor would extend coverage to about 250 m and improve speed estimation accuracy, though at the cost of additional hardware and increased fusion complexity. Radar also produces high data rates, which requires lower end-to-end latency (CPM delay) between the infrastructure unit and the robot than in the current configuration. This delay can be reduced through optimized CPM generation intervals, lighter message payloads, and more efficient scheduling on the V2X channel, which together improve the freshness of perception updates delivered to the robot.

Robot actuation is effective for the tested scenarios, but faster motion would help in borderline cases where the merging vehicle is already close to the intersection while a fast-approaching vehicle is detected on the main road. Lowering the overall response time from detection to gesture execution remains an important direction.

While the system is highly mobile and easily adapts to various intersections without fixed installations, a physical ground-level robot has drawbacks compared to elevated road signs. Sharing the road exposes the robot to vandalism and accidental collisions, especially during active interventions. Additionally, system malfunctions could endanger pedestrians or parked vehicles. Therefore, large-scale deployment requires strict adherence to safety standards and robust fail-safe mechanisms to mitigate these risks.

In addition, future work will focus on extending the concept to more complex traffic scenarios with explicit quantification of robot interventions and its impact on traffic flow. Future work will focus on strengthening the experimental validation through comprehensive quantitative analysis. Furthermore, adaptive decision thresholds that account for traffic density and environmental conditions will be incorporated. Finally, the integration of predictive models for multi-agent interaction will be explored to further enhance the reliability of the moderator in dynamic traffic scenarios.

\section{Conclusion}

This work demonstrated a robotic traffic moderation framework that fuses infrastructure vision and V2X communication to manage NLOS merging scenarios. During the deployment at the FMP, the camera system detected more than seven hundred vehicles and a V2X-equipped VW ID.4 was used to evaluate direct V2X information retrieval, allowing ARI to obtain precise position and speed data. The fused perception enabled the humanoid robot to predict conflicts and physically stop the merging vehicle through a STOP gesture, a handheld stop sign, and a short movement to the center of the lane.

The results show that robotic intervention is effective in mixed traffic environments where many vehicles lack V2X capabilities or where conventional V2X warnings alone cannot guarantee safe driver response. The mobility of ARI and the lightweight sensing unit supports deployment in varied intersection layouts.

\section*{Acknowledgement}
This work was funded by the KIT Future Fields Wild Ideas project "WildRobot", the Excellence Strategy of the German Federal and State Governments and the Helmholtz Programs EDF and ESD. 

This paper is part of the CulturalRoad project, funded by the European Union under grant agreement No. 101147397. Views and opinions expressed are however those of the author(s) only and do not necessarily reflect those of the European Union or the European Climate, Infrastructure and Environment Executive Agency (CINEA). Neither the European Union nor the granting authority can be held responsible for them.

The authors also thank the Future Mobility Network and FMP foundation for providing access to the test site and for their support during the experimental campaign, as well as the staff members who assisted with on site coordination.

\bibliographystyle{IEEEtran} 
\bibliography{ref}

@article{anisha2023automated,
  title={Automated vehicle to vehicle conflict analysis at signalized intersections by camera and LiDAR sensor fusion},
  author={Anisha, Alabi Mehzabin and Abdel-Aty, Mohamed and Abdelraouf, Amr and Islam, Zubayer and Zheng, Ou},
  journal={Transportation research record},
  volume={2677},
  number={5},
  pages={117--132},
  year={2023}
}

@inproceedings{wu2023multi,
  title={A multi-sensor video/lidar system for analyzing intersection safety},
  author={Wu, Aotian and Banerjee, Tania and Chen, Ke and Rangarajan, Anand and Ranka, Sanjay},
  booktitle={2023 IEEE 26th International Conference on Intelligent Transportation Systems (ITSC)},
  pages={1158--1165},
  year={2023}
}

@inproceedings{liu2018cooperation,
  title={Cooperation of V2I/P2I communication and roadside radar perception for the safety of vulnerable road users},
  author={Liu, Weijie and Muramatsu, Shintaro and Okubo, Yoshiyuki},
  booktitle={2018 16th International Conference on Intelligent Transportation Systems Telecommunications (ITST)},
  pages={1--7},
  year={2018},
  organization={IEEE}
}

@article{zhang2021longitudinal,
  title={Longitudinal-scanline-based arterial traffic video analytics with coordinate transformation assisted by 3D infrastructure data},
  author={Zhang, Terry Tianya and Guo, Mengyang and Jin, Peter J and Ge, Yi and Gong, Jie},
  journal={Transportation Research Record},
  volume={2675},
  number={3},
  pages={338--357},
  year={2021}
}

@article{ghaffar2022controlling,
  title={Controlling traffic with humanoid social robot},
  author={Ghaffar, Faisal},
  journal={arXiv preprint arXiv:2204.04240},
  year={2022}
}

@inproceedings{najjar2022towards,
  title={Towards a Smart Robot Model for Traffic Signal Management in Developing Countries},
  author={Najjar, Amro and Prakash, Harisha and Tchappi, Igor and Ndamlabin Mboula, Jean Etienne and Mualla, Yazan},
  booktitle={Proceedings of the 10th International Conference on Human-Agent Interaction},
  pages={333--336},
  year={2022}
}

@inproceedings{mahendran2025design,
  title={Design and Development of Autonomous Drone Traffic Control System},
  author={Mahendran, S and Paul, Stephen and Ashraf, Nissi and Anbarasu, B and Seralathan, S and others},
  booktitle={2025 3rd International Conference on Artificial Intelligence and Machine Learning Applications Theme: Healthcare and Internet of Things (AIMLA)},
  pages={1--6},
  year={2025},
  organization={IEEE}
}

@article{masuduzzaman2022uav,
  title={UAV-based MEC-assisted automated traffic management scheme using blockchain},
  author={Masuduzzaman, Md and Islam, Anik and Sadia, Kazi and Shin, Soo Young},
  journal={Future Generation Computer Systems},
  volume={134},
  pages={256--270},
  year={2022},
  publisher={Elsevier}
}

@article{butilua2022urban,
  title={Urban traffic monitoring and analysis using unmanned aerial vehicles (UAVs): A systematic literature review},
  author={Butil{\u{a}}, Eugen Valentin and Boboc, R{\u{a}}zvan Gabriel},
  journal={Remote Sensing},
  volume={14},
  number={3},
  pages={620},
  year={2022},
  publisher={MDPI}
}

@article{shah2024arow,
  title={AROW: V2X-Based Automated Right-of-Way Algorithm for Cooperative Intersection Management},
  author={Shah, Ghayoor and Tian, Danyang and Moradi-Pari, Ehsan and Fallah, Yaser P},
  journal={IEEE Transactions on Intelligent Transportation Systems},
  volume={25},
  number={9},
  pages={10983--10999},
  year={2024},
  publisher={IEEE}
}

@article{cai2025v,
  title={V-FCW: Vector-based forward collision warning algorithm for curved road conflicts using V2X networks},
  author={Cai, Xiangpeng and Lv, Bowen and Yao, Hanchen and Yang, Ting and Dai, Houde},
  journal={Accident Analysis \& Prevention},
  volume={210},
  pages={107836},
  year={2025},
  publisher={Elsevier}
}

@inproceedings{bazzi2020hardware,
  title={A hardware-in-the-loop evaluation of the impact of the V2X channel on the traffic-safety versus efficiency trade-offs},
  author={Bazzi, Alessandro and Blazek, Thomas and Menarini, Michele and Masini, Barbara M and Zanella, Alberto and Mecklenbr{\"a}uker, Christoph and Ghiaasi, Golsa},
  booktitle={2020 14th European Conference on Antennas and Propagation (EuCAP)},
  pages={1--5},
  year={2020},
  organization={IEEE}
}

@article{gelbal2023cooperative,
  title={Cooperative collision avoidance in a connected vehicle environment},
  author={Gelbal, Sukru Yaren and Zhu, Sheng and Anantharaman, Gokul Arvind and Guvenc, Bilin Aksun and Guvenc, Levent},
  journal={arXiv preprint arXiv:2306.01889},
  year={2023}
}

@techreport{harkey2021impact,
  title={Impact of intersection angle on highway safety},
  author={Harkey, David L and Lan, Bo and Srinivasan, Raghavan and Kumfer, Wesley and Carter, Daniel and others},
  year={2021},
  institution={United States. Federal Highway Administration. Office of Safety Research and~…}
}

@inproceedings{banjade2021vulnerable,
 title={Vulnerable road users safety in infrastructure assisted intelligent transportation 
system},
 author={Banjade, Vesh Raj Sharma and Jha, Satish Chandra and Sivanesan, 
Kathiravetpillai and Baltar, Leonardo Gomes and Sehra, Suman A and Tan, Soo Jin},
 booktitle={2021 IEEE International Smart Cities Conference (ISC2)},
 pages={1--7},
 year={2021},
 organization={IEEE}
}

@article{schlurscheid2024analysis,
  title={An Analysis of Visibility Requirements and Reaction Times of Near-Field Projections},
  author={Schl{\"u}rscheid, Tabea and Stuckert, Alexander and Erkan, Anil and Khanh, Tran Quoc},
  journal={Applied Sciences},
  volume={14},
  number={2},
  pages={872},
  year={2024},
  publisher={MDPI}
}

@inproceedings{khoshkdahan2026TriBand,
  title={TriBand-BEV: Real-Time LiDAR-Only 3D Pedestrian Detection via Height-Aware BEV and High-Resolution Feature Fusion},
  author={Khoshkdahan, Mohammad and Vinel, Alexey},
  booktitle={Proceedings of the 25th International Conference on Autonomous Agents and Multiagent Systems},
  year={2026}
}

@article{arbabzadeh2019hybrid,
  title={A hybrid approach for identifying factors affecting driver reaction time using naturalistic driving data},
  author={Arbabzadeh, Nasim and Jafari, Mohsen and Jalayer, Mohammad and Jiang, Shan and Kharbeche, Mohamed},
  journal={Transportation research part C: emerging technologies},
  volume={100},
  pages={107--124},
  year={2019},
  publisher={Elsevier}
}

@article{delooz2022analysis,
  title={Analysis and evaluation of information redundancy mitigation for v2x collective perception},
  author={Delooz, Quentin and Willecke, Alexander and Garlichs, Keno and Hagau, Andreas-Christian and Wolf, Lars and Vinel, Alexey and Festag, Andreas},
  journal={IEEE access},
  volume={10},
  pages={47076--47093},
  year={2022},
  publisher={IEEE}
}

@inproceedings{lanzer2023interaction,
  title={Interaction effects of pedestrian behavior, smartphone distraction and external communication of automated vehicles on crossing and gaze behavior},
  author={Lanzer, Mirjam and Koniakowsky, Ina and Colley, Mark and Baumann, Martin},
  booktitle={Proceedings of the 2023 CHI Conference on Human Factors in Computing Systems},
  pages={1--18},
  year={2023}
}

@inproceedings{comeca2025social,
  title={Social robots for road safety: Pedestrian crossing assistance use-case},
  author={Comeca, Andy Luis Flores and Masarykova, Nina and Halinkovic, Matej and Galinski, Marek and Laskov, Pavel and Vinel, Alexey},
  booktitle={2025 International Symposium ELMAR},
  pages={53--56},
  year={2025},
  organization={IEEE}
}

@INPROCEEDINGS{comeca2025robots,
  author={Comeca, Andy Luis Flores and Masarykova, Nina and Halinkovic, Matej and Galinski, Marek and Laskov, Pavel and Vinel, Alexey},
  booktitle={2025 IEEE International Automated Vehicle Validation Conference (IAVVC)}, 
  title={Robots for Safer Pedestrian Crossing on Two-Lane Roads}, 
  year={2025},
  pages={1-6},
 }

@inproceedings{khoshkdahan2025beyond,
  title={Beyond Overall Accuracy: Pose-and Occlusion-driven Fairness Analysis in Pedestrian Detection for Autonomous Driving},
  author={Khoshkdahan, Mohammad and Akbari, Arman and Akbari, Arash and Zhang, Xuan},
  booktitle={International Conference on Intelligent Transportation Systems (ITSC)},
  year={2025},
  organization={IEEE}
}

@article{tian2025yolov12,
  title={Yolov12: Attention-centric real-time object detectors},
  author={Tian, Yunjie and Ye, Qixiang and Doermann, David},
  journal={arXiv preprint arXiv:2502.12524},
  year={2025}
}

@inproceedings{ETSI-DENM, 
  title        = {Intelligent Transport Systems (ITS); Vehicular Communications; Basic Set of Applications; Part 3: Specifications of Decentralized Environmental Notification Basic Service}, 
  organization = {ETSI}, 
  number       = {EN 302 637-3}, 
  year         = {2019} 
}

@inproceedings{ETSI-CAM, 
  author       = {ETSI}, 
  title        = {Intelligent Transport Systems (ITS); Vehicular Communications; Basic Set of Applications; Part 2: Specification of Cooperative Awareness Basic Service}, 
  number       = {EN\,302\,637-2 v1.4.1 (2020-05)} 

}

@ inproceedings {ETSI-CPM, 
  title        = {Intelligent Transport Systems (ITS); Vehicular Communications; Collective Perception Service}, 
  organization = {ETSI}, 
  number       = {TS\,103\,324}, 
  year         = {2022} 

}

@inproceedings{khoshkdahan2025fair,
  title={Fair-ped: Fairness evaluation in pedestrian detection using clip},
  author={Khoshkdahan, Mohammad and Kj{\"a}r, Nicholas and Flohr, Fabian B},
  booktitle={2025 IEEE Intelligent Vehicles Symposium (IV)},
  pages={1504--1509},
  year={2025},
  organization={IEEE}
}
\end{document}